\documentclass[conference]{IEEEtran}
\usepackage{amsmath,amssymb}
\usepackage{graphicx}
\usepackage{multirow}
\usepackage{hyperref}
\usepackage{booktabs}
\usepackage{array}
\usepackage{cite}
\usepackage{float}
\usepackage{caption}
\begin{document}

\title{ProvRain: Rain-Adaptive Denoising and Vehicle Detection via MobileNet-UNet and Faster R-CNN}

\author{%
\IEEEauthorblockN{Aswinkumar Varathakumaran}
\IEEEauthorblockA{School of Computer Science Engineering\\
Vellore Institute of Technology\\
Chennai, India\\
aswinkumar.v9805@gmail.com}
\and
\IEEEauthorblockN{Nirmala Paramanandham}
\IEEEauthorblockA{School Of Electronics Engineering \\
Vellore Institute of Technology\\
Chennai, India\\
nirmala.p@vit.ac.in}
}

\maketitle

\begin{abstract}
Provident vehicle detection has a lot of scope in the detection of vehicle during night time. The extraction of features other than the headlamps of vehicles allows us to detect oncoming vehicles before they appear directly on the camera. However, it faces multiple issues especially in the field of night vision, where a lot of noise caused due to weather conditions such as rain or snow as well as camera conditions. This paper focuses on creating a pipeline aimed at dealing with such noise while at the same time maintaining the accuracy of provident vehicular detection. The pipeline in this paper, ProvRain, uses a lightweight MobileNet-U-Net architecture tuned to generalize to robust weather conditions by using the concept of curricula training. A mix of synthetic as well as available data from the PVDN dataset is used for this.  This pipeline is compared to the base Faster RCNN architecture trained on the PVDN dataset to see how much the addition of a denoising architecture helps increase the detection model’s performance in rainy conditions. The system boasts an 8.94\% increase in accuracy and a 10.25\% increase in recall in the detection of vehicles in rainy night time frames. Similarly, the custom MobileNet-U-Net architecture that was trained also shows a 10-15\% improvement in PSNR, a $\sim$5-6\% increase in SSIM, and upto a 67\% reduction in perceptual error (LPIPS) compared to other transformer approaches.
\end{abstract}


\section{Introduction}

Night driving tends to be vastly more treacherous due to low light levels, glare from oncoming headlights, and inclement weather that impairs visibility as well. Daytime driving differs from nighttime driving because while vehicles can often be identified based on the shape, color, and context of the road, at night drivers primarily rely on very subtle cues that the vehicle is approaching. For example, low illumination might provide limited time to react once a driver perceives dim outlines of headlights far away, but, with distractions, divided attention, and delay in recognition, those few seconds may not be enough to avert catastrophe. Advanced driver-assistance systems (ADAS) and autonomous vehicles (AVs) attempt to mimic and improve upon this human ability, by sensing these cues of oncoming vehicles and warning the driver much earlier than any human might be able to react. In order to examine the issue systematically, the Provident Vehicle Detection at Night (PVDN) dataset \cite{ohnemus2020provident} was released, which works differently from a typical object detection dataset in that it focuses on driving at night, and utilizes grayscale scenes with detailed labels for oncoming vehicles, bright light sources, and reflective artifacts. The PVDN dataset aims to introduce early detection from light cues instead of relying on full vehicle identity, which ultimately makes the dataset serve as a benchmark that is more aligned with the way human drivers evaluate threats under low-light driving conditions. 

Existing datasets hold great promise, but the accompanying pipelines tend to ignore the effects of severe weather. Rain in particular poses distinct challenges for improved low-light driving and detection of contrast features, as it creates a visual clutter with streaking and scattering, introduces distortion through raindrops on the lens, and could obscure weak light patterns due to the combination of the motion of vehicles with the rain. Each of these elements can distort the small, yet critical cues that early detection relies on. Though there exist general rain removal, and image enhancement techniques, they have rarely been tailored for this novel challenge.

The proposed solution in this paper, ProvRain, will help to address this use-case with a lightweight, real-time denoising module added to the keypoint detection pipeline while considering the computational limitation of automotive-capable hardware. Nevertheless, beyond pre-processing, a weather-aware curricula training approach is adopted, where models are first trained on clean nighttime images and are then exposed to increasing synthetically and annotatively severe rainy conditions during training. This curricula approach will help model robustness, while maintaining detection models that are capable of discriminating small and critical cues such as the reflection of headlights during heavy rain. The ultimate aim is to achieve provident detection like a human - being able to identify hazards sooner, under the same severe conditions that make nighttime driving overall much more hazardous.

\section{Related Works}

Research surrounding nighttime driving safety has gained momentum in recent years along with the increasing recognition of how much drivers are at an increased risk when driving in dark and/or adverse conditions. To reiterate, although drivers rely on the subtle hints, like the insignificantly faint light on the horizon or slight reflections off of wet asphalt, machine perception systems must also be trained to observe and process these weak signals under unfavorable conditions. And thus, the datasets and architectures developed specifically designed to work in dark, nighttime driving situations have evolved not to apply to broad or general detection during the daytime only, but also anticipate the presence of oncoming vehicles.

The Provident Vehicle Detection at Night (PVDN) dataset \cite{ohnemus2020provident} has gone on to become the predominantly used benchmark for this task. While causal detection has been demonstrated at scale in recent datasets the PVDN captures the environment of nocturnal driving scenarios and emphasizes the explicit annotation of headlights, taillights and reflective artifacts in driving scenarios. All of the architectures evaluated on PVDN by Ivarsson and Zacke (2023) \cite{ivarsson2023} bare a comparison of six different deep learning models. Convolutional neural networks exhibited a relative overall solid ability in detecting other vehicles and consistently outperformed transformer-based detection by a significant margin, with DenseNet recording the highest accuracy 88.10\% accuracy and avg F1 score of 0.83. These examples suggest, that although many researchers have been moving towards transformer architectures for deep learning, the Convolutional neural network still appears to continue to limit its effectiveness to anticipating a localised light pattern in an intrinsically dark, high-noise condition.

Nonetheless, noise and weather continue to be a potentially challenging aspect. Surveys and advances in image deraining \cite{survey_deraining} indicate the necessity of pre-processing for this type of data.  A paper using guided filters showed there was a 12.19\% increase in vehicle counts when using pre-processing under heavy rain \cite{guided_filter_rain}, but this experiment did not truly study the maintenance of light cues at the lower end of the dynamic range.  Motion has been examined under video settings, where FastDVDnet provided real-time denoising while maintaining temporal coherence if any small light artifacts were to be tracked \cite{fastdvdnet}. BM3D and other versions indexed to mean Gaussian-threshold \cite{bm3d} still remain as robust single-frame denoising approaches in low-light conditions. Transformer based models have also gained increasing traction in this concept. SwinIR\cite{swinir} is an image denoiser based on a CNN transformer architecture with windowed self-attention mechanisms. Similarly, Restormer\cite{restormer} is a transformer based architecture with deep learning which uses channel-wise transposed attention and gated depthwise convolutions to remove noise effectively from images. Nevertheless, pre-processing methods still fail to capture weather corruptions when the weather differs from training data. Robustness studies \cite{robustness1,robustness2} repeatedly demonstrate how models are under-performing when faced with previously unseen corruptions. Theoretical frameworks such as curriculum learning have been proposed to address this type of problem, where models can begin with easy tasks and progress towards corrupted or complicated versions. Synthetic data generation has shown promise for addressing dataset scarcity in nighttime vehicle detection for this. Synthetic nighttime data generated with CARLA was utilized in tandem with Efficient Attention GAN (EAGAN) for day-to-night style transfer and producing realistic headlight modeling compared to previously failed GAN approaches \cite{yang2024}. This labeling-free augmentation framework, based on mapping daytime annotations directly to style-transferred nighttime images, led to significant improvements in model performance and significantly higher confidence scores than the baseline models trained on daytime only images. Curriculum augmentation and curriculum smoothing \cite{curriculum_smoothing} have successfully aided the performance of vision models in condition corruptions, but they have not been explored in regards to performance to providential detection.

In conclusion, most of the existing PVDN pipelines adopt a staged runtime consisting of tone mapping or CLAHE preprocessing, lightweight denoising, light-proposal generation (RetinaNet or saliency-based), compact classification, plausibility checks, and temporal tracking. Prior work \cite{prior_modular} shows this modular structure can be effective, but other than showing limited focus on weather noise at low-visibility, there is little prior work on this step. This paper continues this line of work by providing a rainfall-aware denoising step as well as while employing a curriculum parcel-driven training step so that reliability is maintained in the very cases early detection is most critical.

\section{Methodology}
\begin{figure*}
    \centering
    \includegraphics[width=1\linewidth]{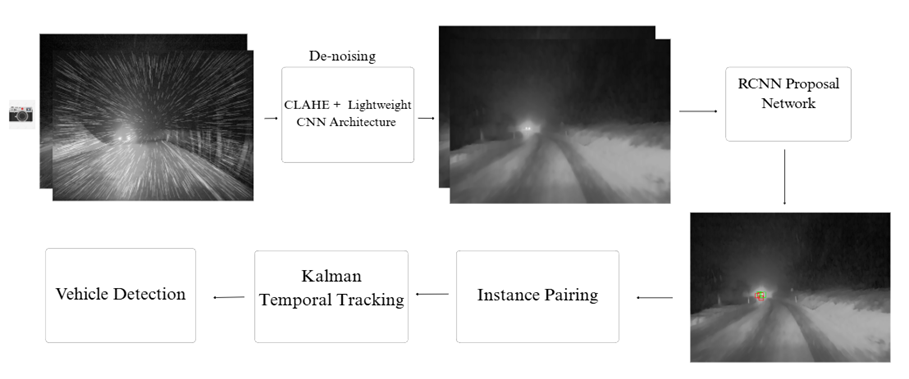}
    \caption{ProvRain Architecture Diagram}
    \label{fig:arc}
\end{figure*}

ProvRain enhances existing provident detection pipelines in two significant ways: (i) weather-aware denoising included in the perception stack, and (ii) a tiered architecture designed to maximize performance in the presence of low-light and adverse-weather conditions, designed specifically to ensure that systems on automotive-grade hardware which may have constraints on computing resources can continue to deploy the system without sacrificing detection fidelity whether it be rain or at night.

\subsection{Preprocessing and Weather-Aware Denoising}

\begin{figure}
    \centering
    \includegraphics[width=1\linewidth]{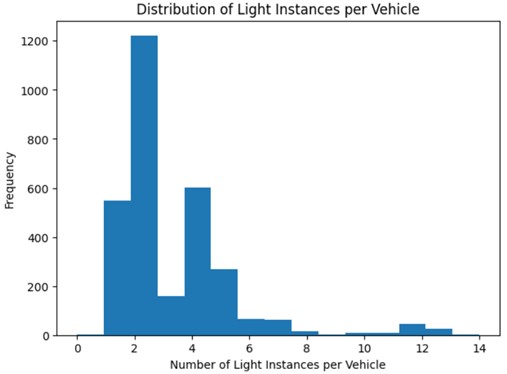}
    \caption{Distribution of Light Instances per Vehicle}
    \label{fig:light_instances}
\end{figure}
Imagery driving under nighttime conditions cleary suffers from a number of noise problems; specifically, sensor noise related to high ISO, artificial noise from stark contrasts of headlights, and environmental weather noise like rain streaks, water droplet reflections, and haze. The first step involves implementing preprocessing to restore visibility at night. 

\begin{table}[ht]
\centering
\caption{Dataset Composition Overview}
\renewcommand{\arraystretch}{1.15} % spacing
\fontsize{9}{11}\selectfont % increase font size
\begin{tabular}{llrrrr}
\toprule
\textbf{Split} & \textbf{Time} &
\textbf{Scenes} &
\textbf{Images} &
\shortstack{\textbf{Vehicle} \\ \textbf{Positions}} &
\textbf{Instances} \\
\midrule

\multirow{2}{*}{\textbf{Train}} 
 & Day   & 113 & 19,078 & 15,403 & 45,765 \\
 & Night & 145 & 25,264 & 26,615 & 72,304 \\
\midrule

\multirow{2}{*}{\textbf{Validation}} 
 & Day   & 20 & 3,898 & 2,602 & 7,244 \\
 & Night & 25 & 4,322 & 3,600 & 12,746 \\
\midrule

\multirow{2}{*}{\textbf{Test}} 
 & Day   & 19 & 3,132 & 3,045 & 9,338 \\
 & Night & 24 & 4,052 & 3,384 & 10,438 \\
\midrule

\multirow{2}{*}{\textbf{Total}} 
 & Day   & 152 & 26,108 & 21,050 & 62,347 \\
 & Night & 194 & 33,638 & 33,599 & 95,488 \\
\midrule

\multicolumn{2}{l}{\textbf{Cumulative}} 
 & \textbf{346} & \textbf{59,746} & \textbf{54,649} & \textbf{157,835} \\
\bottomrule
\end{tabular}
\vspace{-3mm}
\label{tab:dataset_statistics}
\end{table}

The frames are tone-mapped and contrast-enhanced using the basic approach of gamma correction and CLAHE (Contrast Limited Adaptive Histogram Equalization):

\begin{equation}
I_t' = \text{CLAHE}\big(\gamma(I_t)\big)
\label{eq:clahe}
\end{equation}

where $I_t$ is the raw frame, $\gamma (\cdot)$ enhances mid-tone details, and CLAHE restores local contrast in dark regions without amplifying noise excessively.
To train the denoising MobileNet-U-Net backbone for weather artifacts, a progressive noise-injection curricula training approach is used. Because the extreme degradation of rainy images could extreme fit or lose important high-frequency light cues, the architecture learns about clean nighttime images first, then moves to a frame with synthetic rain streaks at a controlled intensity, and then sequentially and iteratively degrades the clean nighttime images until they eventually become the highly degraded frames from the PVDN dataset. This curricula-style training process enables the network to learn a strong mapping of clean $\rightarrow$ moderately degraded $\rightarrow$ heavily degraded images, and improves generalization with real rain. Let the synthetic rain noise at level $k$ be defined as:

\begin{equation}
I_t^{((k))} = J_t + R_t^{((k))} \label{eq:rain}
\end{equation}

where $J_t$ is the clean background frame, and $R_t^{((k))}$ denotes injected rain streaks parameterized by streak density, orientation, and opacity. The denoising network $g_\phi$ is trained to recover $J_t$:

\begin{equation}
J^t = g_\phi(I_t^{(k)})
\end{equation}

with the loss:

\begin{equation}
L_{total}= \lambda_{mse} L_{mse}+\lambda_{percept} (1 - SSIM(I_p,I_t ))+\lambda_{mask} L_{mask} \label{eq:loss}
\end{equation}

where $L_{mask}$ calculates the binary cross entropy loss between the predicted rain mask and the true rain mask whereas $L_{mse}$ calculates the mse between the predicted and target images.

\begin{equation}
L_{mask}= BCE(M_p,M_t )
\end{equation}

\begin{equation}
L_{mask}= MSE(I_p,I_t )
\end{equation}

\subsection{Light Proposal Generation}

\begin{figure}
    \centering
    \includegraphics[width=1\linewidth]{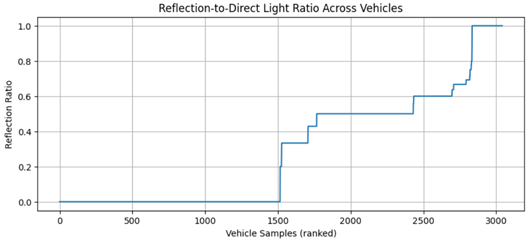}
    \caption{Reflection-to-Direct Light Ratio Across Vehicles}
    \label{fig:light_ratio}
\end{figure}

After post-processing, the system identifies candidate light-like areas. The system primarily identifies lights from the vehicle (and reflections) because they happen earlier and are somewhat more frequent during night time driving conditions. The number of light  instances and the ratio of reflection to direct light sources for the PVDN dataset is seen in Fig~\ref{fig:light_instances} and Fig ~\ref{fig:light_ratio} respectively. Candidate Proposals $P= \{p_i\}$ are generated by a pruned RetinaNet trained with a singular "light-like" class and set with a very low latency,

\begin{equation}
P = \{\, p_i \mid S(I_t') > \tau \,\}
\end{equation}

This allows the system to have a wide coverage of headlight, taillight, and reflection artifacts without rejecting weak signals too early.

\subsection{Proposal Classification}

Each proposal is then categorized into one of three: headlight, taillight, or artifact. A Faster R-CNN classifier (ResNet50) handles each region: 

\begin{equation}
y_i= h_\psi(p_i ) \label{eq:classifier}
\end{equation}

\begin{equation}
{\hat{c}}_i  = P(y_i  \mid p_i)\label{eq:conf}
\end{equation}

Where $c$ is the confidence score. This phase is key to eliminate false positives due to roadside lights, reflective signs, or background clutter.

\subsection{Pairing of instances}

Valid headlights typically come in the form of symmetric pairs. So, pairing rules are used.

\begin{equation}
\{P\}_{pairs} = \{(p_i,p_j )\mid |y_i- y_j |<\epsilon_y, 
 d_{min} \le|x_i- x_j |\le d_{max} \} \label{eq:pair}
\end{equation}

\subsection{Temporal Tracking}

Because headlights exist across several frames, temporal consistency is enforced through Kalman filtering:

\begin{equation}
\{\hat{x}\}_{t\mid t-1} = A \{\hat{x}\}_{t-1\mid t-1} + B u_t \label{eq:kalman1}
\end{equation}

\begin{equation}
P_{t\mid t-1} = A P_{(t-1\mid t-1)} ^T A^T+ Q \label{eq:kalman2}
\end{equation}

with Hungarian algorithm-optimized assignments. Optionally, DeepSORT could be used to obtain richer embedding features to improve identity tracking, and to have smoother confidence curves over time.

\subsection{Decision Logic}

Finally, the decision-making algorithm integrates all cues. A vehicle is flagged as "likely present" if: 
\begin{itemize}
\item It passes the alignment + spacing test, or
\item lights are spreading apart or growing $\rightarrow$ vehicle is coming closer
\item The classifier’s confidence is high enough (higher than required threshold).
\end{itemize}
This is denoted as:

By combining geometric, temporal, and classification signals, the system balances sensitivity and robustness. The entire system architecture is seen in Fig 1.

\begin{table*}[ht]
\centering
\caption{Model Inference Results Comparison}
\label{tab:visual_results}

\begin{tabular}{|>{\centering\arraybackslash}m{3cm}|
                    >{\centering\arraybackslash}m{3cm}|
                    >{\centering\arraybackslash}m{3cm}|
                    >{\centering\arraybackslash}m{3cm}|}
\hline
\textbf{Model Inference Results} &
\textbf{Input Image} &
\textbf{Ground Truth} &
\textbf{Model Output} \\
\hline

SwinIR &
\includegraphics[width=0.16\textwidth]{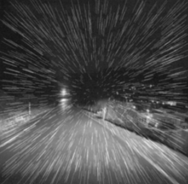} &
\includegraphics[width=0.16\textwidth]{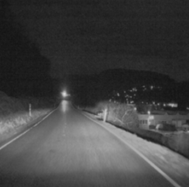} &
\includegraphics[width=0.16\textwidth]{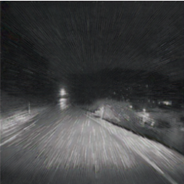} \\
\hline

Restormer &
\includegraphics[width=0.16\textwidth]{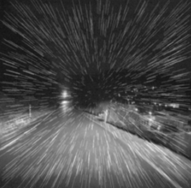} &
\includegraphics[width=0.16\textwidth]{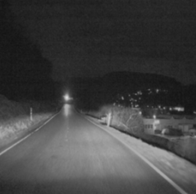} &
\includegraphics[width=0.16\textwidth]{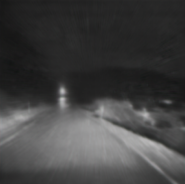} \\
\hline

\textbf{ProvRain (Ours)} &
\includegraphics[width=0.16\textwidth]{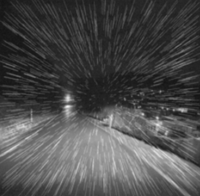} &
\includegraphics[width=0.16\textwidth]{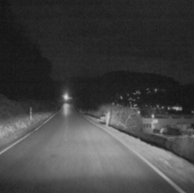} &
\includegraphics[width=0.16\textwidth]{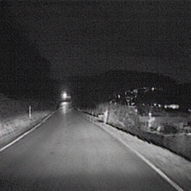} \\
\hline
\end{tabular}
\end{table*}

\section{Data Preprocessing}
\begin{figure}
    \centering
    \includegraphics[width=1\linewidth]{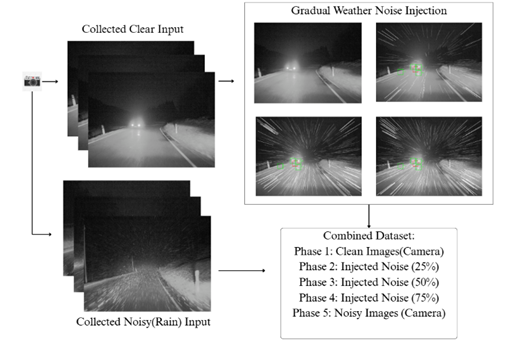}
    \caption{Curricula Training Dataset Preperation}
    \label{fig:curricula}
\end{figure}

The raw dataset has two parts: day and night, with night scenes being the focus of this paper. The taken frames from night scenes were filtered for scenes with rain streaks first. Once the filtering, preprocessing techniques were applied first to get good image normalization ready for inputted data for denoising and detection pipeline purposes. For normalization, the images from the scenes first got resized and normalized to a size that matched the backbone networks (and then intensity scaling to [0,1][0,1][0,1]. To mimic rainy weather degradations, synthetic rain streaks and gaussian blur was added to the clean training images. These streaks were added in a way where they emerge from a vanishing point at the center of the screen, to imitate the rain streaks on the images of the PVDN dataset. Since the dataset’s images consisted of frames taken from a camera moving through the rain, the vanishing point streak approach helped replicate the use case effectively. With a curriculum learning approach, the set was divided into stages of growing levels of noise intensity (Fig~\ref{fig:curricula}):

\begin{enumerate}
\item Stage 1 (clean baseline): clean original frames with photometric correction only.
\item Stage 2 (25\% noise): light, semi-transparent streak overlays were imposed to model drizzle conditions.
\item Stage 3 (50\% noise): Random orientation dense streaks, blur kernels, and light scattering effects were used.
\item Stage 4 (75\% noise): high degradations such as clusters of large streaks, motion blur, and haze-like scattering were introduced.
\item Stage 5 (Mixing of rainy images): The frames in the dataset that contained rain were blended in with the synthetically generated data to facilitate the system's adjustment to real time conditions and render it more robust. 
\end{enumerate}

Training with incrementally increasing noise enabled the denoising network to learn to converge on clean and weakly degraded inputs before acquiring the skill to learn more complex distortions. The augmentation strategy of progressive increases increases robustness while avoiding overfitting to arbitrary artifacts. The synthetic noise was only added to the training data, which reflects more realistic evaluation noise on the training and testing sets.

\section{Observations}

The experiments demonstrated a few useful low-light preprocessing techniques, tone-mapping, gamma correction and CLAHE, that improved visibility of vehicle lights but also wrongfully amplified background noise and reflections with the most noticeable impact coming from video captured under rainy conditions. FastDVDnet\cite{fastdvdnet} originally used as a fast video denoiser retained some static low-light characteristics while denoising. It was ineffective at dynamic denoising and removal of rain streaks, and occasional produced rain streak-like patterns that concealed true light sources and provided degraded input for the downstream detection. From these results it was clear that an off-the-shelf denoiser was not viable for the challenging night-rain context. The SwinIR\cite{swinir} and Restormer\cite{restormer} architectures produced usable results. However, they had their own flaws including improper removal of noise and not preserving the sharpness of the image after denoising respectively. So, a light-weight custom MobileNet-U-Net architecture was taken up which was trained to remove rain noise. modules (the light proposal generator and the classifier) increased in reliability.  Paired with the Faster R-CNN architecture, the classifier now correctly differentiated the light artifacts in scenes previously misclassified under heavy rain conditions.

It was also observed that using light proposals based on simple geometric heuristics together with temporal smoothing through Kalman filtering was effective in generating stable tracks of vehicles. Even without any explicit depth information provided, symmetry and motion coherence were enforced to filter out spurious detections from reflections and artifact noise stemming from rapidly disappearing objects. With curriculum trained custom Mobilenet-U-Net removing rain noise, the system also could maintain robust tracks over time, which reflected in detections being earlier and being more trusted. All in all, moving away from generic denoisers towards task-specific lightweight Mobilenet-U-Net with curriculum learning was integral to obtaining strong performance under the night-rain conditions without sacrificing real-time efficiency.

\section{Results And Inference}

The methodology in this paper is compared to the original Faster RCNN architecture used on the PVDN dataset. The custom MobileNet-U-Net denoising network, attached to the Faster RCNN pipeline was qualitatively measured against the base model on metrics including recall, accuracy, early warning success and inference time (Table~\ref{tab:visual_results}).
ProvRain was trained mainly on Kaggle. The P100 GPU with 16GB VRAM was used, enabling efficient experimentation and faster convergence across platforms. The light custom MobileNet-U-Net denoiser demonstrates the greatest improvement with the suggested pipeline. The night rain images with no denoising had distracting streaks and noise that made automatic detection unstable; the light proposal generator had frequent false positive detections and the classifier reassigned mistaken reflections and rain debris to car lights. Early warnings were therefore less reliable.

When the MobileNet-U-Net denoiser was introduced, the images that noisy with rain were more easily processed and essential light areas were identifiable. This further improved the proposal generator enabling the classifier to detect headlights and taillights with higher precision. It proved to be much more efficient than other denoising techniques including the SwinIR and Restormer which are equally performative model, however did not capture and retain image sharpness as well after removal of noise(Table~\ref{tab:visual_results}). The peak signal noise ratio (PSNR) score, structural similarity Index (SSIM) and mean absolute error (L1 Loss) were used to compare the models (Table~\ref{tab:denoise}).

\begin{table}[t]
\centering
\caption{Comparison of Denoising Techniques}
\label{tab:denoise}
\renewcommand{\arraystretch}{1.2} % Optional: better row height
\begin{tabular}{|l|c|c|c|}
\hline
\textbf{Metric} & \textbf{SwinIR} & \textbf{Restormer} & \textbf{ProvRain Denoiser (Ours)} \\
\hline
\textbf{PSNR (dB)} & 31.45 & 32.72 & 36.24 \\
\hline
\textbf{SSIM} & 0.8874 & 0.8961 & 0.941 \\
\hline
\textbf{L1 Loss} & 0.0108 & 0.0099 & 0.0096 \\
\hline
\textbf{MSE} & 0.00072 & 0.00051 & 0.000238 \\
\hline
\textbf{RMSE} & 0.0268 & 0.0226 & 0.0186 \\
\hline
\textbf{MAE} & 0.0132 & 0.0120 & 0.0111 \\
\hline
\textbf{LPIPS} & 0.1623 & 0.1389 & 0.1264 \\
\hline
\end{tabular}
\end{table}

\begin{figure}
    \centering
    \includegraphics[width=1\linewidth]{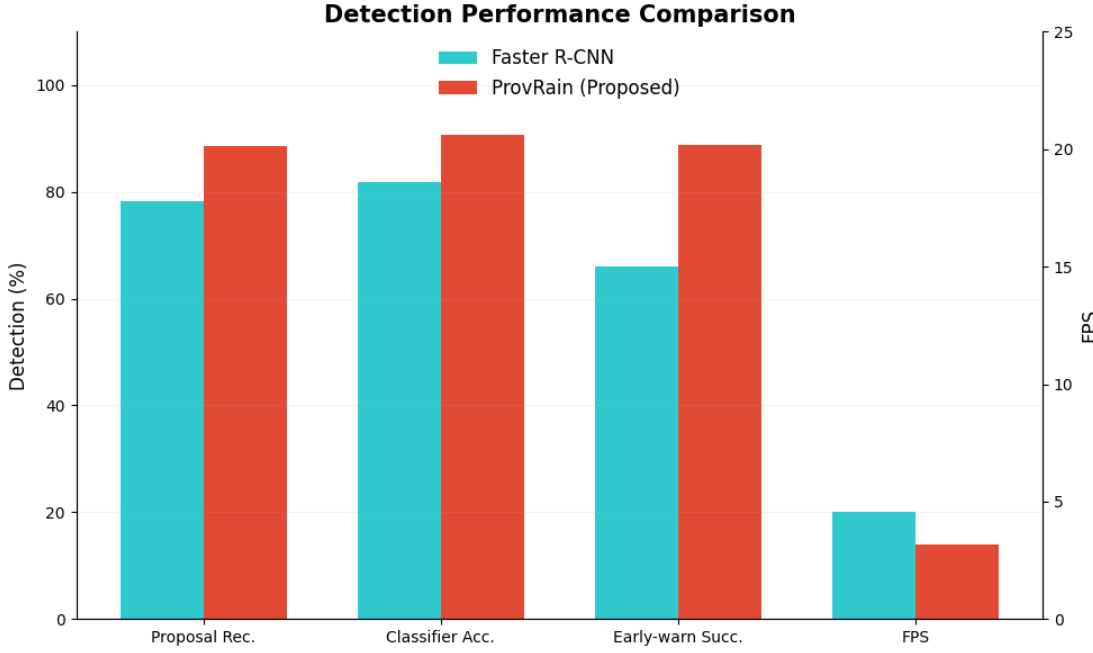}
    \caption{Detection Performance Comparison}
    \label{fig:detection_performance}
\end{figure}

Quantitatively, the custom Mobilenet-U-Net denoiser increased detection stability and confidence generally with little to no extra cost. By comparison of the raw images to the denoised images, it was visually verified that car lights were still visible in the denoised outputs while rain streaks and other artifacts were suppressed, for the most part. These results demonstrate that for reliable nighttime car detection in rain, task specific lightweight denoising through curriculum learning proves to be advantageous, leading to an uptick in performance from using raw images alone, without sacrificing the speed of the entire pipeline for the real-time deployment (Table\ref{tab:results}).

\begin{table}[t]
\centering
\caption{Inference Results of Proposed Methodology}
\label{tab:results}
\renewcommand{\arraystretch}{1.15}

\begin{tabular}{|
    p{2.1cm}|
    >{\centering\arraybackslash}p{2.25cm}|
    >{\centering\arraybackslash}p{2.25cm}|
}
\hline
\textbf{Metric} &
\textbf{Faster R-CNN Without Denoising} &
\textbf{ProvRain Pipeline (Ours)} \\
\hline
Proposal Recall (\%) & 78.28 & 88.53 \\
\hline
Classifier Accuracy (\%) & 81.74 & 90.68 \\
\hline
Early-warning Success (\%) & 65.92 & 88.72 \\
\hline
Avg frames per second & 20 & 14 \\
\hline
\end{tabular}
\end{table}

\section{Conclusions and Future work}

The methodology in this paper, ProvRain, presents a lightweight, real-time pipeline for the providential detection of vehicles in severe raining conditions at night. By combining the image preprocessing, the task-specific MobileNet-U-Net with curriculum learning, efficient light proposal generation, classification, and motion tracking, the system is capable of detecting headlights and taillights reliably whilst filtering out rain noise and rain artefacts. The experiments on the PVDN dataset illustrated that the curriculum-trained denoiser can significantly increase detection stability and track persistence, as well as the early warning success when contrasting to using the raw images.

In the future, this pipeline can be improved by in a multitude of ways. Generalizing the curriculum learning framework to also be used in additional weather scenarios such as fog or snow could further increase generalizability to different night-driving conditions. Looking into adaptive confidence levels and novel pairing heuristics could lead to even less false positives, and even earlier warnings. Finally, linking the pipeline to utilise multi camera set-ups or monocular depth cues from motion could potentially allow the pipeline to perform more accurate motion estimation without increased computational cost.

\end{document}